\def\year{2020}\relax
\documentclass[letterpaper]{article} % DO NOT CHANGE THIS
\usepackage{aaai20}  % DO NOT CHANGE THIS
\usepackage{times}  % DO NOT CHANGE THIS
\usepackage{helvet} % DO NOT CHANGE THIS
\usepackage{courier}  % DO NOT CHANGE THIS
\usepackage[hyphens]{url}  % DO NOT CHANGE THIS
\usepackage{graphicx} % DO NOT CHANGE THIS
\usepackage{graphicx}  % DO NOT CHANGE THIS
\usepackage{booktabs}       % professional-quality tables
\usepackage{amsfonts}       % blackboard math symbols
\usepackage{nicefrac}       % compact symbols for 1/2, etc.
\usepackage{microtype}      % microtypography
\usepackage{subcaption}
\usepackage{algorithm}
\usepackage{algorithmic}
\usepackage{multirow}
\usepackage{amsmath}
\usepackage{verbatim}

\title{Learning to Auto Weight: Entirely Data-driven and Highly Efficient  Weighting Framework}
\author{
Zhenmao Li,\textsuperscript{\rm 1}
Yichao Wu,\textsuperscript{\rm 1}
Ken Chen,\textsuperscript{\rm 1}\\
\Large
\textbf{Yudong Wu,}\textsuperscript{\rm 1}
\textbf{Shunfeng Zhou,}\textsuperscript{\rm 1}
\textbf{Jiaheng Liu,}\textsuperscript{\rm 2}
\textbf{Junjie Yan}\textsuperscript{\rm 1}\\
\textsuperscript{\rm 1}SenseTime\hspace{2pt}
\textsuperscript{\rm 2}BUAA \\
{\{lizhenmao,wuyichiao,wuyudong,zhoushunfeng,yanjunjie\}@sensetime.com}\\
{kenchen1024@gmail.com}\hspace{4pt}
{liujiaheng@buaa.edu.cn}
}

\begin{document}

\maketitle

%%%%%%%%% ABSTRACT
\begin{abstract}
  Example weighting algorithm is an effective solution to the training bias problem,
however,
most previous typical methods are usually limited to human knowledge and
require laborious tuning of hyperparameters.
In this paper,
we propose a novel example weighting framework called \textit{Learning to Auto Weight} (LAW).
The proposed framework finds step-dependent weighting policies adaptively, and can be jointly trained
with target networks
without any assumptions or prior knowledge about the dataset.
It consists of three key components:
\textit{Stage-based Searching Strategy (3SM)}
is adopted to shrink the huge searching space in a complete training process;
\textit{Duplicate Network Reward (DNR)}
gives more accurate supervision
by removing randomness during the searching process;
\textit{Full Data Update (FDU)}
further improves the updating efficiency.
Experimental results demonstrate the superiority of weighting policy explored by LAW over standard training pipeline.
Compared with baselines,
LAW can find a better weighting schedule which achieves much more superior accuracy on both biased CIFAR and ImageNet.
\end{abstract}

%-------------------------------------------------------------------------

\section{Introduction}
Although the quantity of training samples is critical for current state-of-the-art deep neural networks (DNNs),
the quality of data also has significant impacts on exerting the powerful capacity of DNNs on various tasks.
For supervised learning,
it is a common hypothesis that both training and test examples are drawn i.i.d. from the same distribution.
However,
during practical training,
this assumption is not always valid,
therefore,
the training bias problems, mainly including label noise and class imbalance, are encountered frequently.

It is widely known that the sample weighting algorithm is an effective solution to the training bias problem.
Although common techniques such as cost-sensitive learning~\cite{lin2017focal} and curriculum learning~\cite{bengio2009curriculum,kumar2010self}
demonstrate the effectiveness of example reweighting,
they are usually predefined for specific tasks with prior knowledge and
require laborious tuning of hyperparameters.
To alleviate this problem adaptively,
a method of learning to auto-weight is probably effective,
which searches weighting strategies for picking more valuable samples or filtering harmful samples.
However, there are some inherently severe challenges if we want to make the searing process work well.
\textbf{The first challenge} is the huge searching space caused by numerous iterations during the training process.
Our objective is to search step dependent weighting strategies for making the accuracy on validation datasets as high as possible.
Suppose a complete training process involves thousands of steps $N$,
the number of possible weights for a sample is $w_n$,
the batch size is $B$ (typical 128),
so the number of possible weights of a batch data in one step is $a_n=w_n^B$,
the searching space of weights is $a_n^N=w_n^{BN}$, which is an enormous number for searching a good strategy.
\textbf{The second one} is the randomness which is an implicit but harmful problem for searching good weighting strategies.
The randomness can derive from different data combination,
the random augmentation,
different initialization of parameters, etc.
Thus, given a strategy model for weighting strategies,
using accuracies on validation datasets to update the strategy model may cause the searching process to fail easily.
\textbf{Last but not least}, collecting the training samples for learning to auto weight is nontrivial.
In practice,
to obtain credible feedback signals,
we need to conduct complete training processes,
so that huge numbers of networks must be trained to convergence.
Therefore, the process of searching a weighting strategy is time-consuming with low efficiency.
\\
\indent Based on the above analysis,
in this paper, we propose a novel example weighting strategies searching framework to learn weighting strategies from data adaptively,
which is modeled by the strategy model.
For the first challenge,
we simplify the searching process and divide the training process into a
small number of stages $N^\prime$ (typical 20) consisting of successive iterations,
so that the time steps for searching can be significantly limited to the number of stages $N^\prime \ll N$.
We call this method \textit{\textbf{Stage-based Searching Strategy Method (3SM)}},
which can shrink the searching space and reduce time costs significantly.
Meanwhile, to solve the second challenge, we design a novel feedback signal measurement,
called \textit{\textbf{Duplicate Networks Reward (DNR)}},
where a reference network is added to generate accuracy difference on the validation dataset as the feedback signal to remove the randomness.
In this way,
if we get a higher accuracy or lower accuracy,
the searching algorithm could focus on the quality of the weighting strategy, and the strategy model could be updated for better ones.
Besides, to raise the efficiency in the last challenge,
we utilize a data buffer to cache the samples for updating the strategy model,
and make full of all the data in the buffer to optimize the strategy model for numbers of epochs.
We call this updating method \textit{\textbf{Full Data Update (FDU)}},
which can improve the efficiency of updating the strategy model significantly and accelerate the weighting strategies searching.
Experimental results demonstrate the superiority of weighting policy explored by LAW over standard training pipeline.
Especially,
compared with baselines,
LAW can find a better weighting schedule which achieves much more superior accuracy in the noisy or imbalance CIFAR and ImageNet dataset.

Our contributions are listed as follows:
\begin{itemize}
    \item [1)]
    We propose a novel example weighting strategy searching framework called LAW,
    which can learn weighting policy from data adaptively.
    LAW can find good sample weighting schedules that achieve higher accuracy in the contaminated and imbalance datasets
    without any extra information about the label noises.
    \item [2)]
    We propose Stage-based Searching Strategy Method (3SM) to shrink the huge searching space in a complete training process.
    \item [3)]
    We design novel Duplicate Networks Reward (DNR) that removes the data randomness and make the process of searching weighting strategies more effectively.
    \item [4)]
    We propose Full Data Update (FDU) to make full use of all the data in the buffer to make the searching strategy process more efficient.
\end{itemize}

%----------------------------------------------------------------------------
\section{Related Work}
\textbf{Curriculum Learning:} Inspired by that humans learn much better when putting the examples in a meaningful order (like from easy level to difficult level),
\cite{bengio2009curriculum} formalizes a training strategy called curriculum learning which promotes learning with examples of increasing difficulty.
This idea has been empirically verified and applied in a variety of areas~\cite{kumar2010self,lee2011learning,supancic2013self,jiang2014easy,graves2017automated}.
Self-paced method~\cite{kumar2010self} defines the curriculum by considering the easy items with small losses in early stages and add items with large losses in the later stages.
\cite{peng2019accelerating} builds a more efficient batch selection method based on typicality sampling,
where the typicality is estimated by the density of each sample.
\cite{jiang2014self} formalizes the curriculum with preference to both easy and diverse samples.
\cite{alain2015variance} reduces gradient variance by the sampling proposal proportional to the L2-norm of the gradient.
%The core problem is the measure of item's hardness which is usually determined by heuristic rules.
Curriculums in the existing literature are usually determined by heuristic rules and thus require laborious tuning of hyperparameters.

\textbf{Weighting:}
The practice of weighting each training example has been well
investigated in the previous studies.
Weighting algorithms mainly solve two kinds of problems:
label noise and class imbalance.
If models can converge to the optimal solution on the training set with coarse labels,
there could be large performance gaps on the test set.
This phenomenon has also been explained in \cite{zhang2016understanding,neyshabur2017exploring,arpit2017closer}.
Various regularization terms on
the example weights have been proposed to prevent
overfitting to on corrupted labels~\cite{ma2017self,jiang2015self}.
Recently,
Jiang et al. propose MentorNet~\cite{jiang2017mentornet},
which provides a weighting scheme for StudentNet to focus on the sample whose label is probably correct.
However, to acquire a proper MentorNet,
it is necessary to give extra information such as the correct labels on a dataset during training.
On the other hand, the class imbalance is usually caused by the cost and difficulty in collecting rarely seen classes.
Kahn and Marshall \cite{kahn1953methods} propose importance sampling which assigns
weights to samples to match one distribution to
another.
Lin et al.~\cite{lin2017focal} propose focal loss to address the class imbalance by adding a soft weighting scheme that emphasizes harder examples.
Other techniques such as cost-sensitive weighting~\cite{khan2018cost} are also useful for class imbalance problems.
Previous methods usually require prior-knowledge to determine a specified weighting mechanism,
the performance will deteriorate if we cannot get accurate descriptions of the dataset.
To learn from the data,
Ren et al. \cite{ren2018learning} propose a novel meta-learning algorithm that
learns to assign weights to training examples based on their gradient directions.

\begin{figure*}[]
  \centering
  \includegraphics[width=0.8\linewidth]{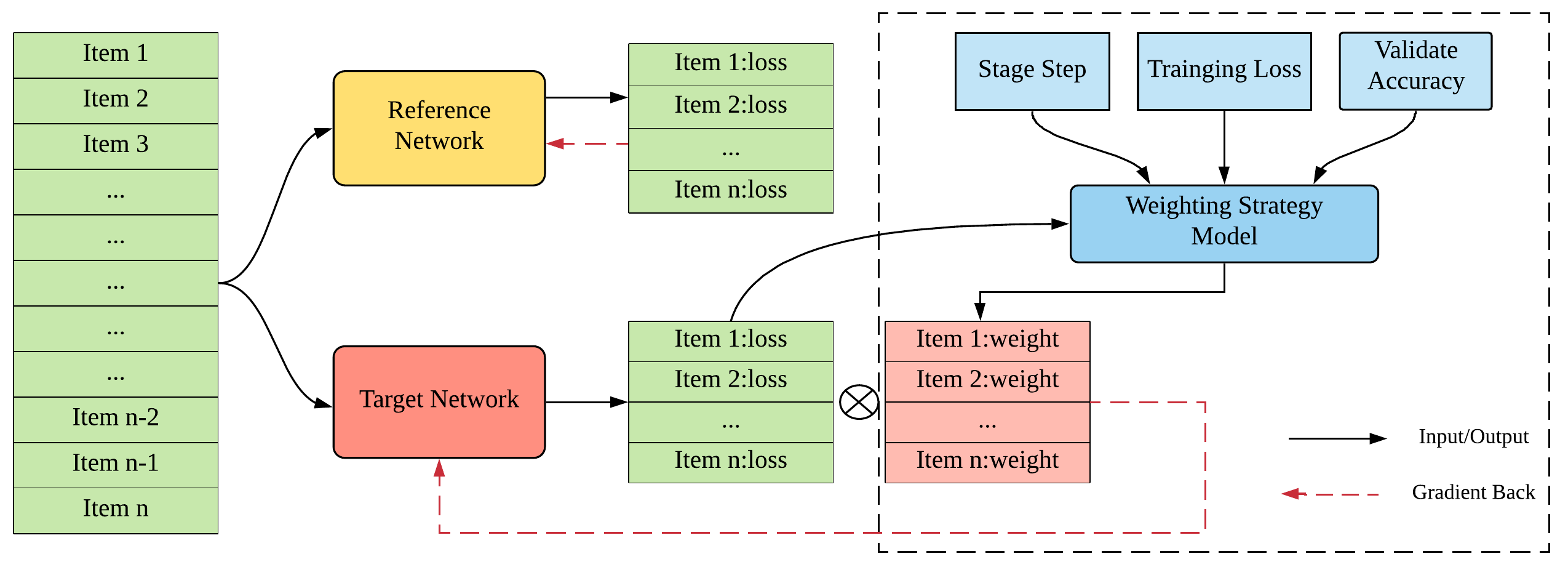}
  \caption{The sample weighting framework of LAW.}
\label{framework}
\end{figure*}

%------------------------------------------------------------------------
\section{Learning to Auto Weight}
In this section,
we first demonstrate the formulation of learning to auto weight as a problem of searching
the best strategy to pick valuable samples for training.
It's also a bilevel optimization problem of training a classification model
and a strategy model for weighting samples.
Second, we explain the framework to solve the optimization problem.
In the end, we describe the learning to search the best weighting strategy in detail.
\subsection{Problem Formulation}
\label{pf}
Unlike the standard SGD training process,
which treats every sample equally in all batches and all training stages,
LAW tries to find the best strategy to weight samples in different batches
and steps for better accuracy on validation datasets.
In this paper,
we model the strategy as a sample weighting function $\mathcal{K}(f,\theta)$
parameterized by $\theta$, where the $f$ is the feature of one sample.
In different training steps $t=1,2,3, ... T$,
there are different weighting functions denoted by $\mathcal{K}_t(f,\theta_t)$ of different weighting strategies.
Given an train dataset $\mathcal{X}_{train}=\{(x_i,y_i)|i=1,2,3 ...N_{train}\}$
and a validation dataset $\mathcal{X}_{val}=\{(x_i,y_i)|i=1,2,3 ...N_{val}\}$,
we need to train a network $\mathcal{M}(\cdot,w)$ parameterized by $w$ for the classification or other tasks.
The purpose of LAW is to find a weighting strategy making a training network achieve better accuracy in all steps,
and this can be realized by maximizing the cumulative
validation accuracy associated with the network $\mathcal{M}(\cdot,w)$,
which is trained to minimize corresponding losses.
Thus, the objective of LAW is:
\begin{align}\label{object}
& \max_{\theta} \mathcal{J}(\theta) = \sum_{t}acc(w_t^*) \\
& acc(w_{t}^*) = \frac{1}{N_{val}}\sum_{(\hat{x},\hat{y})\in {\mathcal{X}_{val}}} \delta(\mathcal{M}(\hat{x},w_{t}^*),\hat{y}) \\
{\rm{s.t.}} \quad &{w_t^*}  = \mathop{\arg}\mathop{\min} \limits_{w}\sum_{(x,y)\in{\mathcal{X}_{train}}}\frac{{\mathcal{K}_t(f,\theta_t)\mathcal{L}}(\mathcal{M}(x,w),y)}{N_{train}}, \nonumber
\end{align}
where the $\delta$ is an impulse function,
whose value is 1 when $\mathcal{M}(\hat{x},w_{t}^*)$ is equal to $\hat{y}$ and 0 otherwise,
$\mathcal{L}$ denotes the loss function.
%This refers to a typical sequential decision making problem,
There are different weighting strategies for different training steps.
For example, in early steps,
the network may favor easy samples while in later stages,
hard examples may need be considered much more.
To tackle this problem,
we sample weighting strategies according to the $\mathcal{K}_t(f,\theta_t)$,
then train classification networks to obtain the best models $\mathcal{M}(\cdot,w^*)$.
Next, the corresponding accuracies provide signals to optimize the $\mathcal{K}_t(f,\theta_t)$.
Thus,
the weighting strategy would be found under these two optimization processes interactively.

\subsection{Weighting Framework}
Our framework is illustrated in Figure~\ref{framework},
where the left half side constitutes the training procedure on classification networks
and right side constitutes the weighting procedure by the strategy model,
named Weighting Strategy Model defined as $\mathcal{K}(f,\theta)$ in \ref{pf}.
In every training step,
we sample a batch of data consisting of n data items and feed them to two networks with identical architectures,
Reference Network and Target Network.
Reference Network is trained by a general training procedure without any weighting strategy and Target Network is trained by the
weighting strategy from the Weighting Strategy Model.
We collect the training state of Target Network,
including Stage Step, Training Loss and Validation Accuracy (the details will be elaborated later) with some features of data items like losses.
Then Weighting Strategy Model outputs weights for every data item.
Thus, a weighted loss mean is calculated by an element multiplication between the losses and corresponding weights.
In the end, the weighted loss mean is utilized to optimize Target Network,
which is different from the optimization of Reference Network.

\begin{figure}[h]
  \centering
  \includegraphics[width=\linewidth]{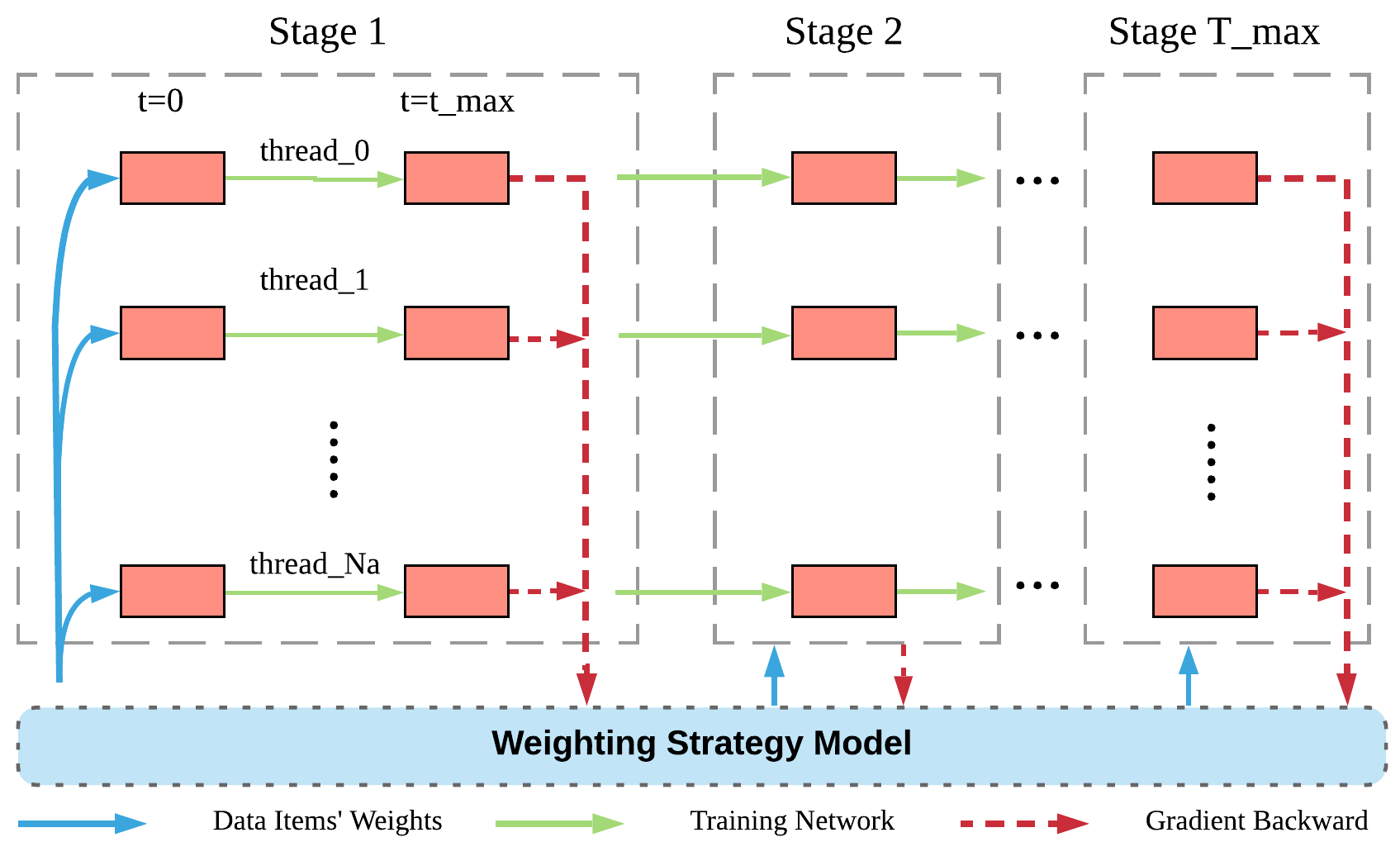}
  \caption{Illustration of stage-based weighting strategy searching. We conduct numbers of threads in parallel and synchronize the gradients of the weighting strategy model.}
\label{stage}
\end{figure}

\textbf{Stage-based Searching Strategy Method (3SM):} For a standard SGD training process,
one iteration consists of a forward and backward propagation based on the input mini-batch of samples,
and the whole training process usually contains large numbers of successive iterations before getting the final model.
In this process,
the weights of feeding samples may have influences on the accuracy in the end,
especially for biased training datasets.
However,
on account of the thousands of iteration steps,
it's tricky and inefficiency to search a strategy based on one iteration step.
Therefore,
we uniformly divide the training process to a small number of stages,
where one weighting strategy keeps unchanged for all iteration steps in one stage until the start of the next stage.
To clarify this,
we use $T=1,2,3 ... T_{max}$ to denote the stage,
and $t=1,2,3 ...t_{max}$ to denote iteration steps of training a network inside one stage.
So the number of total iteration steps is $T_{max}\times{t_{max}}$.
Weighting Strategy Model outputs weights to weight all samples from $t=1$ to $t=(t_{max}-1)$,
where there is no update for the strategy model.
While at $t=t_{max}$,
we calculate feedback signals and update the strategy model by the Full Data Update (the details will be elaborated later).
The 3SM is illustrated in Figure~\ref{stage}.

In the first step $t=1$,
we collect informations of the Target Network as training phase descriptor $s_T$ in every stage,
such as current training stage $T$, smoothed historical training loss $l_{smooth}$,
smoothed historical validating accuracy $acc_{smooth}$.
To improve the expression capacity,
we embed the number of the current training stage to a $d$ dimension vector: $T \rightarrow \textbf{e}_T \in R^d$, which is initialized using Gaussian initialization and is optimized as a part of Weighting Strategy Model in every searching step.
When the current training stage is $T$,
we define an intermediate network as $\mu(s_T|\theta^\mu)$ parameterized by $\theta^\mu$
and the output of $\mu(s_T|\theta^\mu)$ is the parameter $\theta_T$ in $\mathcal{K}_T(f,\theta_T)$,
where $f$ is items' features descriptor.
The features we used are listed as follows:\\
\textbf{\textit{Training Loss}}:
One practicable descriptor is the training loss,
which is frequently utilized in the curriculum learning, hard example mining and self-paced methods.
The loss can be used to describe the difficulty level of a sample and
sometimes it would be helpful to drop out the samples with large losses when there are outliers.\\
\textbf{\textit{Entropy}}:
For classification tasks,
the entropy of predicted probabilities demonstrates the hardness of the input samples.
Hard examples tend to have large entropy while samples with small and stable entropy are probably easy.\\
\textbf{\textit{Density}}: The density of one item reveals how informative or typical the sample is,
which can be used to measure the importance of samples.
Obviously,
samples with large density should be paid more attention to.
%The density of features could be calculated and saved beforehand by Gaussian kernel density algorithm~\cite{peng2019accelerating}.
To make it simple,
we calculate similarity matrices using samples' logits $\textbf{Lo}$ defined as $\textbf{Lo}^T\textbf{Lo}$ in one batch,
and for each sample, we average it's similarities with other samples to approximate the density.\\
\textbf{\textit{Label}}: We can also use the label on account of that the label information
would help to remove some bias in the dataset like class imbalance.

In addition,
we normalize the features to make the learning process more stable and the learned strategy more general for other datasets and networks.
Once the features are extracted, the weights of items are defined as:
\begin{align}\label{weight_function}
&\mathcal{K}_T(f,\theta_T)=1+\tanh(\theta_Tf+b),\\
\label{theta}&\theta_T=\mu(s_T|\theta^\mu),\\
\label{s}&s_T=[\textbf{e}_T, l_{smooth}, acc_{smooth}],\\
&f=[loss,entropy,density,label],
\end{align}
%Specifically,
%the Weight Strategy Model receives $s_T$ in each stage,
%and outputs a vector as the $\theta_T$ to decide each sample's weight.
For $t=0,1,2 ... t_{max}$ in stage $T$,
the gradients of the loss weighted by $\mathcal{K}_T(f,\theta_T)$ will be propagated back
to update parameters of the Target Network $\mathcal{M}(\cdot,w)$:
\begin{equation}\label{update_w}
w_{t} = w_{t-1} - \eta \nabla_{w}[\mathcal{K}_{T} \times \mathcal{L}_b(w_{t-1},x_b,y_b)],
\end{equation}
where $\eta$ is the learning rate for the network parameters, $b$ is the batch size, $\mathcal{L}_b$ denotes the batched loss which takes three inputs: the current network parameters $w_{t-1}$ and a mini-batched data $x_b$, $y_b$.

\subsection{Learning to Search Strategy}
Considering that the validation accuracy is non-differentiable with respect to $\theta$ in $\mathcal{K}(f,\theta)$,
it is a tricky problem to calculate the gradient of validation accuracy with respect to $\theta$.
To address this optimization problem,
we utilize the method in DDPG \cite{lillicrap2015continuous} to solve the searching problem approximately.
Specifically,
we add an extra network defined as $Q(s_T,\theta_T|\theta^Q)$ parameterized by $\theta^Q$ to estimate the objective in Equation \ref{object},
where the $\theta_T$ is defined same as $\mathcal{K}_T(f,\theta_T)$.
Thus, if the $Q(s_T,\theta_T|\theta^Q)$ estimates the objective precisely, we can choose $\theta_T$ to maximum the objective,
that is, improving the accuracy on validation datasets.

\textbf{Duplicate Networks Reward (DNR):}
Randomness is an implicit but harmful problem for searching good weighting strategies.
The randomness can be from the SGD optimizer,
different data combination in one batch,
data orders between batches,
random augmentation of inputs,
some random operation like dropping out in a network architecture,
different initialization of parameters,
and numerical precision in hardware et al.
The randomness above could disturb weighting strategy searching because of too many factors that influence the accuracy in the end,
so it's hard for the learning algorithm to decide what makes the accuracy higher.
Thus,
the feedback signal (it's also called reward) for updating the strategy model,
must be designed to remove the randomness so that the learning method can apply credits to those better weighting strategies.
Therefore,
In each episode,
we train two networks,
one for searching strategies called Target Network and the other for comparison called Reference Network.
There are some issues should be noted:
%In addition, there are some consistences in DRN,
The first is that the two network architectures is completely identical;
The second is that the initialization of parameters is completely identical as we copy one network's parameters to another network directly;
The third is that the data inputs at every step are completely identical to remove data randomness.
In iteration steps updating the strategy model,
we calculate accuracies of the Target Network and Reference Network on validation datasets and the reward is defined as the difference of accuracy between them.
We call the reward from two identical networks Duplicate Networks Reward (DNR).
In this way,
if we get a higher accuracy or lower accuracy,
the searching algorithm could put enough credit on better weighting strategies so that the strategy model could be updated forward better ones.
What's more,
Considering that the reward is not important in early stages,
we add different weights from weakness to mightiness on rewards at different stages:
\begin{align}
\label{reward_weight}
&reward = r_w*(acc_{target}-acc_{reference}),\\
&r_w=exp(k*\frac{ce}{ne})*s,
\end{align}
where $ce$ is current epoch,
$ne$ is total number of epochs of training process and $k,s$ is the scale adjustment rate.

\textbf{Full Data Update (FDU):}
Collecting the training samples for learning to auto weight is nontrivial.
In practice,
only if conducting complete training processes, we can obtain credible reward signals.
In this way,
huge numbers of networks must be trained to convergence.
Hence, for updating the strategy model,
we utilize a buffer to cache transitions data~\cite{lillicrap2015continuous},
which is defined by a tuple $(s_T,\theta_T, r_T,s_{T+1},Done)$,
where $T$ is the number of stage, $s_T$ is defined in Equation~\ref{s}, $r_T$ is the reward in stage $T$,
and $Done$ is a bool flag indicating whether the training procedure is finished or not.
Since the time step is based on the stage,
we calculate the reward and update the strategy model only if the iteration step is $t_{max}$ in one stage.
Instead of sampling a batch of transition data in one time step,
we utilize all the data in the buffer to update the strategy model for numbers of epochs,
so that the strategy model can take full advantage of the cached data to learn useful knowledge.
We call this update method Full Data Update (FDU).
FDU can improve the efficiency of updating the strategy model significantly and accelerate the weighting strategies searching.
As illustrated in Figure~\ref{stage},
Multiple classification networks and the strategy models are trained in different threads in parallel simultaneously
and the number of threads is $N_{a}$.
The parameters of the strategy models are shared among all the threads,
and we synchronize gradients when updating the parameters of the strategy model in the last step in one stage.
On the contrary, the parameters of the classification networks are optimized independently without sharing with others.

Given $\mu(s|\theta^\mu)$ parameterized by $\theta^\mu$ and $Q(s,\theta|\theta^Q)$ parameterized by $\theta^Q$,
we have:
\begin{align}\label{critic_u}
&\nabla_{\theta^Q}Q=\frac{1}{N_a}\sum_{n=1}^{N_a}\frac{1}{B}\sum_{i=1}^{B}\nabla_{\theta^Q}L_Q(Q(s_i,\theta_i)),\\
&L_Q(Q(s_i,\theta_i))=(y_i-Q(s_i,\theta_i))^2,\\
&y_i=r_i+\gamma Q(s_{i+1},\mu(s_{i+1}|\theta^{\mu})),\\
\label{actor_u} &\nabla_{\theta^\mu}\mu
\approx\frac{1}{N_a}\sum_{n=1}^{N_a}\frac{1}{B}\sum_{i=1}^{B}\nabla_{\theta}Q(s_i,\theta|\theta^Q)|_{\theta=\mu(s_i)}\nabla_{\theta^\mu}\mu(s_i|\theta^\mu),
\end{align}
where $B$ is batch size.
The algorithm to update the strategy model is illustrated in Algorithm~\ref{alg:u}.
The algorithm details of LAW are list in the Algorithm~\ref{alg:LAW}.

\begin{algorithm}[h]
	\caption{LAW: Update the strategy model}
	\label{alg:u}
	\begin{algorithmic}
		\REQUIRE
		\STATE the buffer $R$, number of epochs $En$, batch size $B$;\\
		\FOR{$e=1,En$}
		\STATE Shuffle all the data in $R$
		\FOR{each $B$ transitions data}
		\STATE Update $Q(s,\theta|\theta^Q)$ one step as in Equation~\ref{critic_u}
		\STATE Update $\mu(s|\theta^\mu)$ one step as in Equation~\ref{actor_u}
		\ENDFOR
		\ENDFOR
		
	\end{algorithmic}
\end{algorithm}

\begin{algorithm}[h]
	\caption{LAW:Learning to auto weight on one thread}
	\label{alg:LAW}
	\begin{algorithmic}
		\REQUIRE
		Training data $D$, batch size $B$, number of total training procedures L, number of total steps in one training procedure $N$, number of steps in one stage $K$;
		
		\STATE
		Randomly initialize $\theta^\mu$ in $\mu(s|\theta^\mu)$ and $\theta^Q$ in $Q(s,a|\theta^Q)$;\\
		%Copy the weights to the target network $\mu^\prime$ and $Q^\prime$: $\theta^{\mu^\prime} \leftarrow \theta^\mu$, $\theta^{Q^\prime} \leftarrow \theta^Q$\\
		%Initialize replay buffer $R$;
		
		\FOR {$episode = 1,L$}
		\STATE Random initialize $w_t$ in Target Network $M_t$
		\STATE Copy $w_t$ to $w_r$ in Reference Network $M^r$: $w^r \leftarrow
		w_t$\\
		%Initialize a random process $P$ for action exploration
		\FOR {$t=1,N$}
		\IF{$t\mod K == 0$}
		\STATE Calculate $\theta_t$ according to Equation~\ref{theta}
		\ELSE
		\STATE $\theta_t=\theta_{t-1}$
		\ENDIF
		\STATE Get data items' weights according to Equation~\ref{weight_function}
		\STATE Optimize Reference Network one step
		\STATE Optimize Target Network one step according to Equation~\ref{update_w}

		\IF{$t\mod K == 0$}
		\STATE Compute the reward by Equation~\ref{reward_weight}\\
		\IF{$t \geq K$}
		\STATE Cache the tuple $(s_{t-K},\theta_t,r_t,s_{t}, Done)$ in the buffer\\
		\ENDIF
		\STATE Update the strategy model according to Algorithm~\ref{alg:u}
		\ENDIF
		\ENDFOR
		\ENDFOR
		\ENSURE The strategy model
	\end{algorithmic}
\end{algorithm}

%------------------------------------------------------------------------
\section{Experiments}

\subsection{Implementation Details}
\label{imp}
We demonstrate the effectiveness of LAW on image classification dataset CIFAR-10,
CIFAR-100~\cite{krizhevsky2009learning},
and ImageNet~\cite{deng2009imagenet}.

\textbf{CIFAR}: CIFAR-10 and CIFAR-100 consist of 50,000 training
and 10,000 validation color images of 32$\times$32 resolution with 10 classes and 100 classes receptively.
They are balanced datasets where each class holds the same number of images.
To search the weighting strategy,
we use a part of the training dataset like 20,000 for training and 5,000 for validation.
While for testing the strategy,
we use all samples.
General pre-processing steps are used in training,
such as zero-padding with 4 pixels,
random crops with size 32$\times$32,
random flips and standardizing the data.
All the networks are trained to convergence from scratch utilizing SGD optimizer with a batch-size of 128.
The weight decay is set to $2e-5$ and the momentum is set to 0.9.
The initial learning rate is 0.1, and then the learning rate is divided by 10 when the stage is 10,13 and 16.
The total number of training epochs is 200,
thus the classification network would be harmed by biased datasets if no weighting strategy was applied.

\textbf{ImageNet}: ImageNet dataset contains 1.28 million training images and 50,000 validation images with 1,000 classes.
We also sample a small part of the training dataset for searching the weighting strategy.
In detail,
the sampled images coverage all the classes,
and contain 100 images per class.
That is,
the total number of the sampled dataset is 100,000.
The process of training ImageNet follows the convention for state-of-art ImageNet models.
Concretely,
the pre-processing includes random flipping,
random size crop to 224$\times$224 and standardization with mean channel subtraction.
We use synchronous SGD with a momentum rate of 0.9 and $2e-5$ weight decay.
The step learning rate scheme that decreases the initial learning rate 0.1 by a factor 0.1 every 30 epoch is utilized.
We also add a warmup stage of 2 epochs.
The batch size is set to 1024 and the number of total training epochs is 100.

For Weight Strategy Model,
the $\mu(s|\theta^\mu)$ and $Q(s,\theta|\theta^Q)$ are modeled using MLP with 4 layers,
which is optimized by Adam~\cite{kingma2014adam} with learning rates $10^{-5}$ and $10^{-4}$ respectively.
We utilize 8 threads for searching weighting strategies in parallel to make this process more efficiently.
Multiple threads can boost the searching process and 8 is an empirical value.
When the $N_a$ is 4,
the computational complexity is lower, but the searching process slows down compared with 8 threads.
On the other hand,
16 threads raise the computational complexity and the searching process will be time-consuming.
For every complete training,
we divide the total number of training epochs uniformly to 20 stages.
What's more,
we set a warmup process considering that weighting strategies of early training stages do not have much impact on the accuracy in the end.
That is,
we don't utilize any weighting strategy in early training stages but train networks in a general procedure which treats every sample equally.
After the warmup,
the strategy model is optimized at every stage to search proper strategies to focus on important samples by weighting.

%All the experiments are implemented on the platform of PyTorch~\cite{paszke2017automatic}.
%It should be mentioned that our codes are attached in the supplementary material
%and will be released in \url{https://github.com/lzm10214066/LearningtoAutoWeight}.

\subsection{Results of effects on noisy labels compatibility}
For noisy datasets,
the label of each image in the training dataset is independently changed to a uniform random class with probability $p$,
where $p$ is set to 0.4.
The labels of images in the validation dataset remain unchanged for evaluation.

\begin{table}[b]
	\centering
	\caption{The top-1 accuracy on CIFAR with noise rate $p=0.4$ of LAW for different networks.}
	\label{cifar_tabel}
	\begin{tabular}{|c|c|c|}
		\hline
		Method & CIFAR-10 & CIFAR-100                                         \\ \hline
		ResNet-18(Base)        & 79.44                 & 55.18                \\ \hline
		ResNet-18(LAW)         & \textbf{86.33}        & \textbf{61.27}       \\ \hline
		WRN-28-10(Base)        & 76.77                 & 53.11                \\ \hline
		WRN-28-10(LAW)         & \textbf{89.73}        & \textbf{68.23}       \\ \hline
		ResNet-101(Base)       & 75.57                 & 52.20                \\ \hline
		ResNet-101(LAW)        & \textbf{88.90}        & \textbf{68.01}       \\ \hline
	\end{tabular}
\end{table}

\textbf{CIFAR Results:} On the CIFAR-10 and CIFAR-100,
we evaluate two ResNet~\cite{he2016deep} networks,
one small 18-layer ResNet, one large 101-layer ResNet and a medium WRN-28-10~\cite{zagoruyko2016wide} network.
%All the networks are trained to convergence from scratch utilizing SGD optimizer with a batch-size of 128.
%The weight decay is set to $2e-5$ and the momentum is set to 0.9.
%The initial learning rate is 0.1, then the step learning rate is divided by 10 when the stage is 10,13 and 16.
%The total number of training epochs is 200,
%thus the classification network would be harmed by biased datasets if no weighting strategy was applied.

%We conduct general classification task on the noisy dataset with corrupted labels.
%Theoretically,
%CIFAR is a balance dataset,
%where there is no obvious difference between images in one class,
%the corroded labels could add significant distinction between samples.
%Thus, our LAW can filter the noisy samples out.
And we illustrate the test accuracy of our LAW for different networks on CIFAR in Table~\ref{cifar_tabel}.
The baseline is the base model that treats every sample equally in every batch.
Compared with baseline models,
our LAW exhibits a significant accuracy improvements for all the networks,
which reveals that our LAW filters the noise data effectively.
%On CIFAR-10,
%LAW achieves over 6 percentage points for ResNet-18, 13 percentage points for WRN-28-10 and 12 percentage points for ResNet-101.
%On CIFAR-100,
%LAW achieves over 5 percentage points for ResNet-18, 15 percentage points for WRN-28-10 and 16 percentage points for ResNet-101.
%Obviously, the improvement of accuracy for heavy architectures is larger than the small networks.
%One of the reason is that the heavy networks is much easier to fit the noise data,
%and our LAW filters the noise data effectively.

To show the effectiveness on noisy datasets further,
we also perform experiments compared with popular methods with WRN-28-10 utilizing human knowledge or meta-learning,
including Self-paced~\cite{kumar2010self},
MentorNet~\cite{jiang2017mentornet},
L2RW~\cite{ren2018learning},
Focal loss~\cite{lin2017focal},
and CO-teaching\cite{han2018co}.
All methods are evaluated under the same setting described in Implementation Details except
the number of total epochs is 120 and the learning rate is divided by 10 when the stage is 18 and 19.
As shown in Table \ref{compare_tabel},
our LAW obviously outperforms any other methods.
%In addition,
%It's worth noting that our LAW dose not need any prior distribution about the noise,
%which is tricky to get any information of the noise in a dataset.
%while the MentorNet~\cite{jiang2017mentornet} need labels of noisy data points and
%perform supervised training to remove noisy data points.
%However,
%in practice,
%it's tricky to get any information of the noise in a dataset.

\newcommand{\tabincell}[2]{
	\begin{tabular}{@{}#1@{}}#2\end{tabular}
}

\begin{table}[t]
	\centering
	\caption{The top-1 accuracy on noisy CIFAR with a 0.4 noise fraction for WRN-28-10 compared with other methods.}
	\label{compare_tabel}
	\begin{tabular}{|c|c|c|c|}
		\hline
		Method & \tabincell{c}{CIFAR-\\10-Noisy(\%)}  & \tabincell{c}{CIFAR-\\100-Noisy(\%)} \\ \cline{2-3}
		\hline
		Self-paced       & 86.21                  & 46.23                               \\ \hline
		MentorNet        & 87.56                  & 65.56                               \\ \hline
		L2RW             & 87.04                  & 62.45                               \\ \hline
		Focal Loss       & 74.12                  & 50.71                               \\ \hline
		Co-teaching      & 74.84                  & 46.45                               \\ \hline
		LAW          & \textbf{89.73}         & \textbf{68.23}                          \\ \hline
	\end{tabular}
\end{table}

\textbf{ImageNet Results:} 	We perform our LAW method on several popular networks on ImageNets: ResNet-18, ResNet-34, ResNet-50,
and Mobilenetv2~\cite{sandler2018mobilenetv2} to demonstrate the effectiveness on large dataset.
As the same with CIFAR,
we also construct a noisy dataset with probability 0.4.
These networks including small,
medium and heavy architectures, are evaluated to illustrate the generalization of our LAW method.
Table~\ref{imagenet_tabel} shows the top-1 test accuracy on validation set.
All the experiments are conducted following the same setting and all the networks are trained from scratch.
Considering the high cost of training process on ImageNet,
we sample a small part of the dataset and train the weighting model only on ResNet-18,
and train the small dataset in one GPU for searching the weighting strategies,
then we transfer the learned weighting strategies to other networks.
To test the learned strategy, we train the networks on ImageNet using synchronous SGD.
As can be seen in the table,
on the noisy ImageNet,
our LAW improves the accuracy obviously.
Similar to the results on noisy CIFAR,
we also find that it can achieve more significant improvement when the network is heavier.

\begin{table}[t]
\centering
\caption{The top-1 accuracy of on noisy ImageNet with noise rate $p=0.4$ of our LAW for different networks.}
\label{imagenet_tabel}
\begin{tabular}{|c|c|}
\hline
Method  & Noisy ImageNet(\%)          \\ \hline
ResNet-18(Base)     & 64.7            \\ \hline
ResNet-18(LAW)      & \textbf{65.2}   \\ \hline
ResNet-34(Base)     & 65.4            \\ \hline
ResNet-34(LAW)      & \textbf{66.5}   \\ \hline
ResNet-50(Base)     & 71.9            \\ \hline
ResNet-50(LAW)      & \textbf{73.7}   \\ \hline
MobileNet-V2(Base)  & 59.3            \\ \hline
MobileNet-V2(LAW)   & \textbf{60.3}   \\ \hline
\end{tabular}
\end{table}

\subsection{Effects on imbalance data}
To evaluate effects of our LAW on the imbalance data,
On CIFAR-10,
we make an imbalance dataset by random discarding $96\%$ of samples with the label of $0$ and $1$,
while keeping the others the same as origin.
That is, for the classes of $0$ and $1$,
we only utilize 200 images on CIFAR-10 to train a network while 5,000 for other classes.
On CIFAR-100,
for the classes of 0 to 9,
we utilize 50 images on CIFAR-100 to train a network while 500 for other classes.
As for networks,
we train VGG-19~\cite{simonyan2014very} and ResNet-18 with the same setting described in Implementation Details above.
It can be seen in the Table~\ref{imbalance_tabel} that the weighting strategies explored by LAW can deal with this problem well.

\begin{table}[t]
\centering
\caption{The top-1 accuracy on imbalance CIFAR of our LAW}
\label{imbalance_tabel}
\begin{tabular}{|c|c|c|}
\hline
Method & \tabincell{c}{CIFAR-10-\\Imbalance(\%)}    & \tabincell{c}{CIFAR-100-\\Imbalance(\%)}      \\ \hline
Vgg-19(Base)         & 84.88                 & 59.33                \\ \hline
Vgg-19(LAW)         & \textbf{85.33}        & \textbf{60.20}       \\ \hline
ResNet-18(Base)      & 85.93                 & 60.24                \\ \hline
ResNet-18(LAW)      & \textbf{88.44}        & \textbf{61.94}       \\ \hline
\end{tabular}
\end{table}

To compare with other methods,
we use Long-Tailed CIFAR dataset~\cite{cui2019class} and set the imbalance rate to 100,
that is, the number of the largest class is 100 times the number of the smallest class.
Other popular methods include
Focal loss~\cite{lin2017focal},
Class-Balanced~\cite{cui2019class},
L2RW~\cite{ren2018learning},
and we conduct experiments with ResNet-32 following the training schedule above except
the number of total epochs is 100 and the learning rate is divided by 10 when the stage is 14 and 18.

\begin{table}[htp]
\centering
\caption{The top-1 accuracy of ResNet-32 on imbalance CIFAR compared with other methods}
\label{imbalance_tabel_compare}
\begin{tabular}{|c|c|c|}
\hline
Method & \tabincell{c}{CIFAR-10-\\Imbalance(\%)}    & \tabincell{c}{CIFAR-100-\\Imbalance(\%)}      \\ \hline
Focal Loss            & 71.34                 & 39.41                 \\ \hline
Class-Balanced        & 74.6                  & 40.16                 \\ \hline
L2RW                  & 74.25                 & 41.23                 \\ \hline
LAW                   & \textbf{76.34}        & \textbf{43.61}        \\ \hline
\end{tabular}
\end{table}

\subsection{Analysis}
In this section,
we perform several analyses to illustrate the effectiveness of learned strategies by our LAW.

\begin{figure}[h]
	\begin{center}
		\begin{subfigure}[t]{0.48\linewidth}
			\centering
			\includegraphics[width=1.03\linewidth]{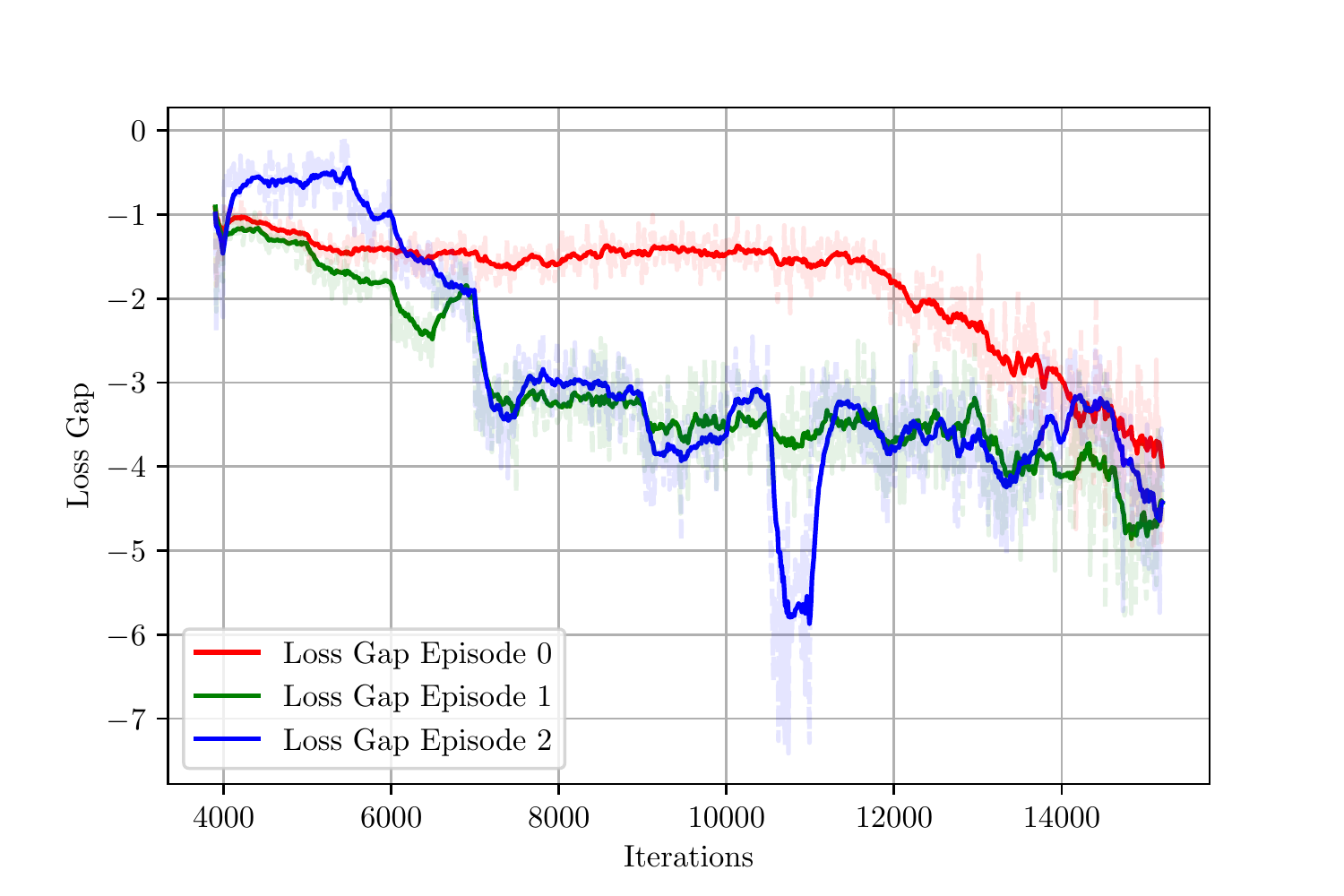}
			\caption{CIFAR10-Noise}
			\label{noise_loss_gap_cifar10}
		\end{subfigure}
		\begin{subfigure}[t]{0.48\linewidth}
			\centering
			\includegraphics[width=1.03\linewidth]{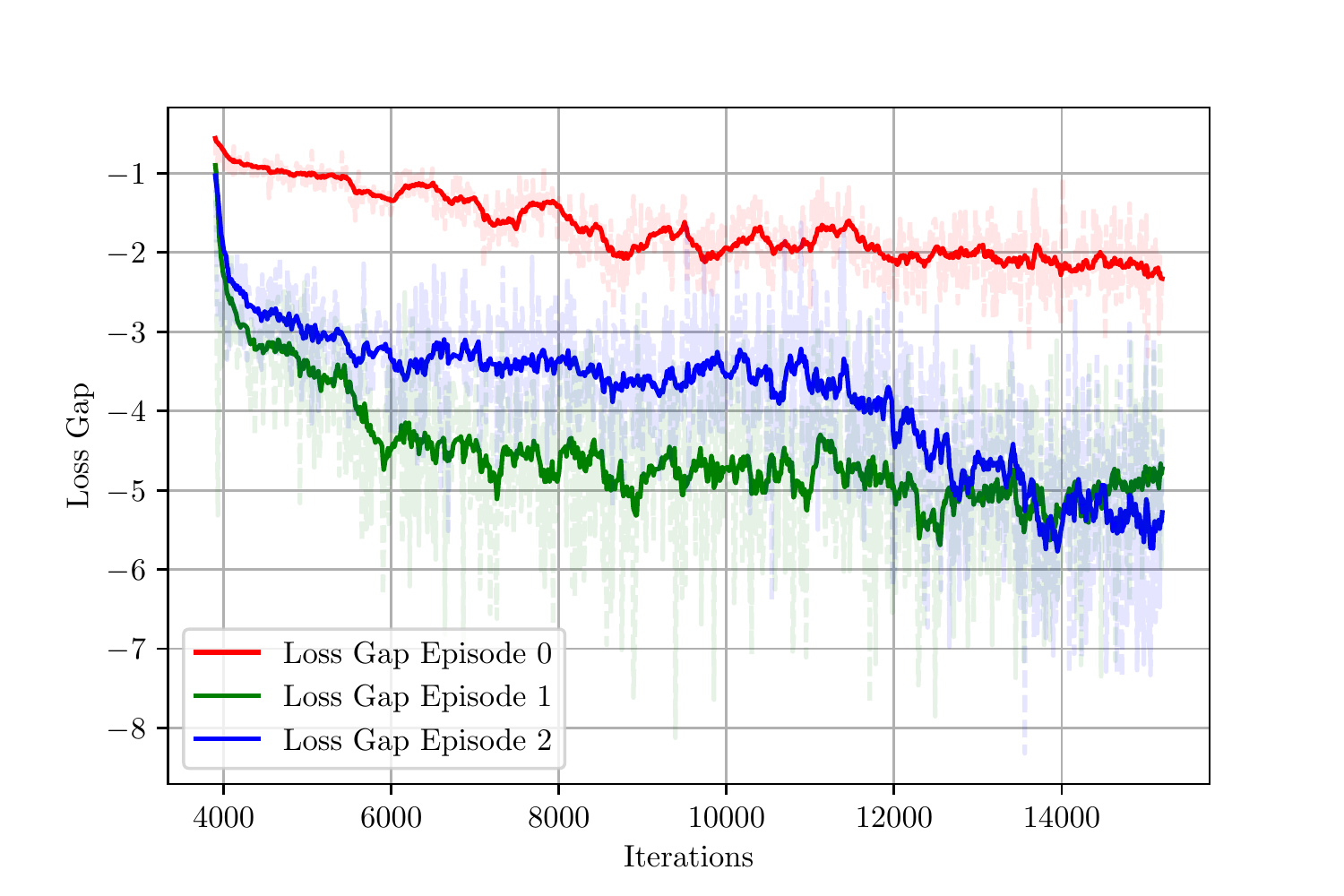}
			\caption{CIFAR100-Noise}
			\label{noise_loss_gap_cifar100}
		\end{subfigure}
		\begin{subfigure}[t]{0.48\linewidth}
			\centering
			\includegraphics[width=1.07\linewidth]{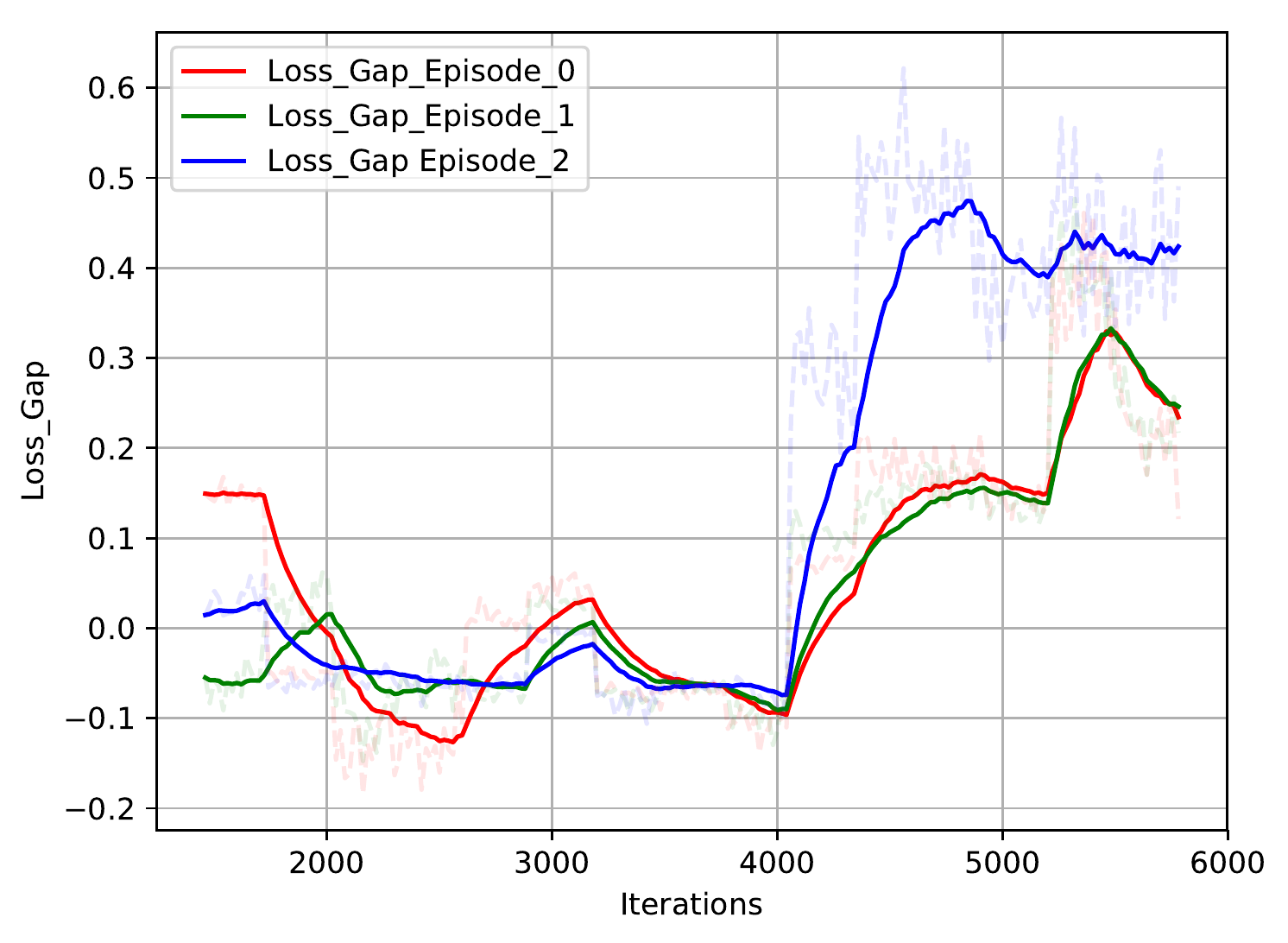}
			\caption{ImageNet-Clean}
			\label{loss_gap_imagenet}
		\end{subfigure}
		\begin{subfigure}[t]{0.48\linewidth}
			\centering
			\includegraphics[width=1.03\linewidth]{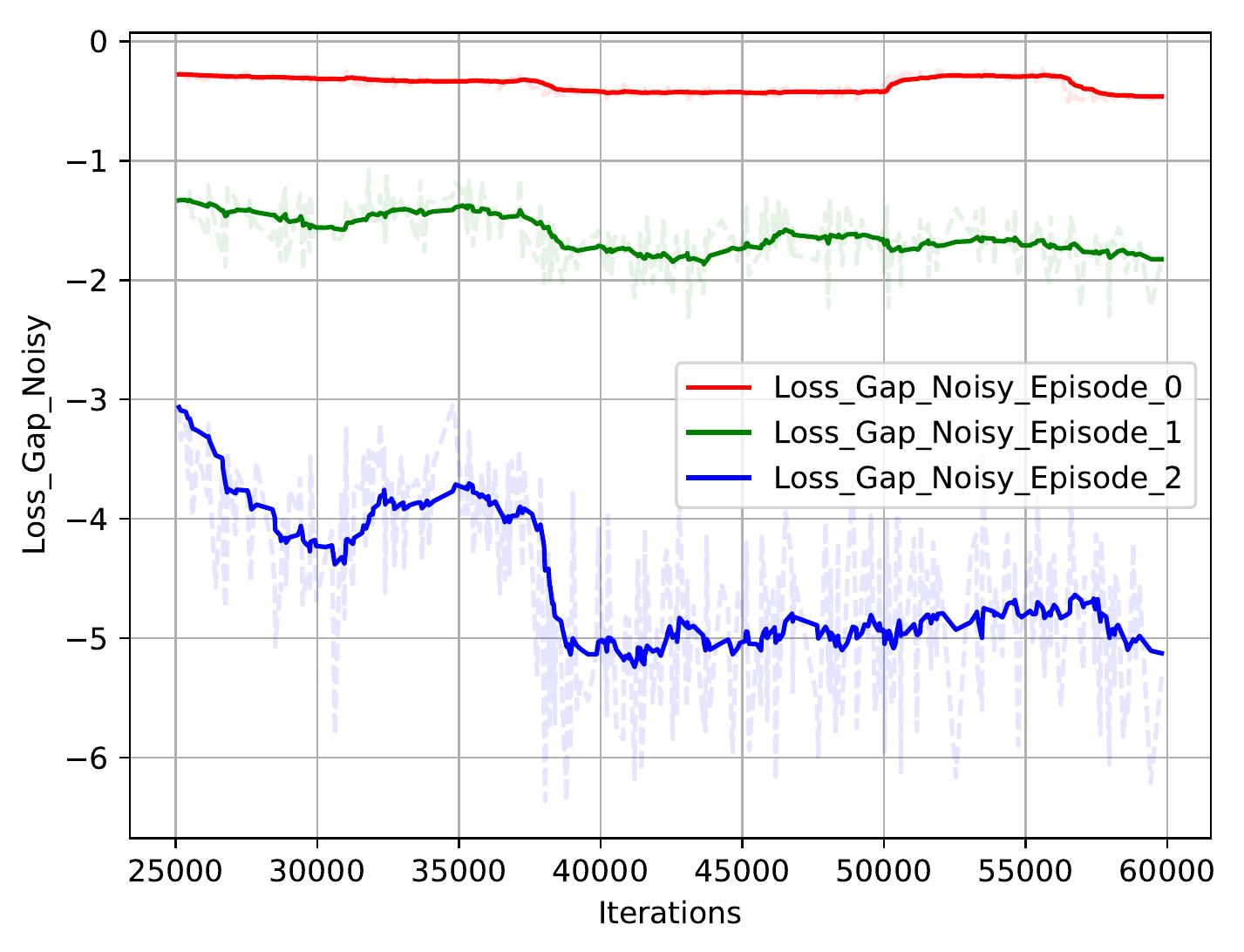}
			\caption{ImageNet-Noise}
			\label{noise_loss_gap_imagenet}
		\end{subfigure}
	\end{center}
	\caption{The loss mean gap between target network and reference network in random three episodes.}
	\label{loss_gap}
\end{figure}

The curves of losses gap are shown in Figure~\ref{loss_gap},
where the loss gap is defined as the mean of items' losses in one batch between the network trained
with the learned weighting strategies and the network trained without any weighting strategy.
For noisy datasets,
tendencies of loss gaps are amazingly consistent for all datasets.
Figure~\ref{noise_loss_gap_cifar10},
Figure~\ref{noise_loss_gap_cifar100} and Figure~\ref{noise_loss_gap_imagenet} illustrate loss gaps in noisy CIFAR-10,
CIFAR-100 and ImageNet respectively.
Apparently, the loss gaps are all under zero,
which demonstrates that the learned weighting
strategy from LAW can distinguish those data instances with corrupted labels and reduces the weights of them.
For both CIFAR and ImageNet,
the final accuracy of target networks is significantly higher than that of reference networks.
It shows that LAW can find a much effective weighting schedule to find noisy data instances and filter them.
To make the comparison more clearly,
we also perform our LAW on the clean ImageNet and draw the loss gap in Figure~\ref{loss_gap_imagenet}.
Figure~\ref{loss_gap_imagenet} demonstrates that the most of loss gap is near the 0 value
and above 0 value in the later stage,
which is contrary to the results on noisy ImageNet.
That is,
our LAW can find the noisy samples effectively and filter them in the early stages so that the damage from noise can be eliminated.

\begin{figure}[h]
	\begin{center}
		\begin{subfigure}[t]{0.48\linewidth}
			\centering
			\includegraphics[width=1.05\linewidth]{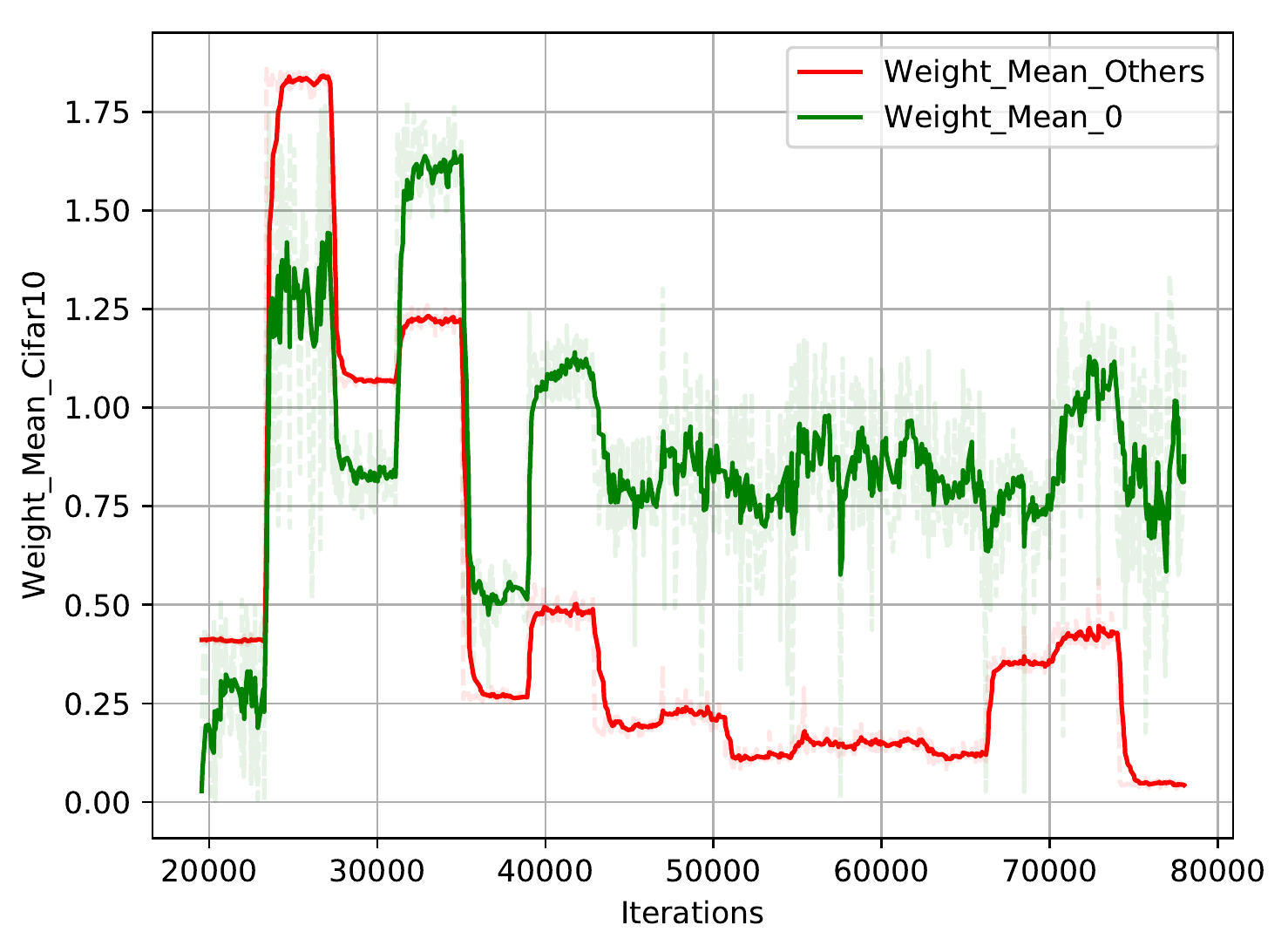}
			\caption{CIFAR10-Imbalance}
			\label{imbalance_weight_cifar10}
		\end{subfigure}
		\begin{subfigure}[t]{0.48\linewidth}
			\centering
			\includegraphics[width=1.05\linewidth]{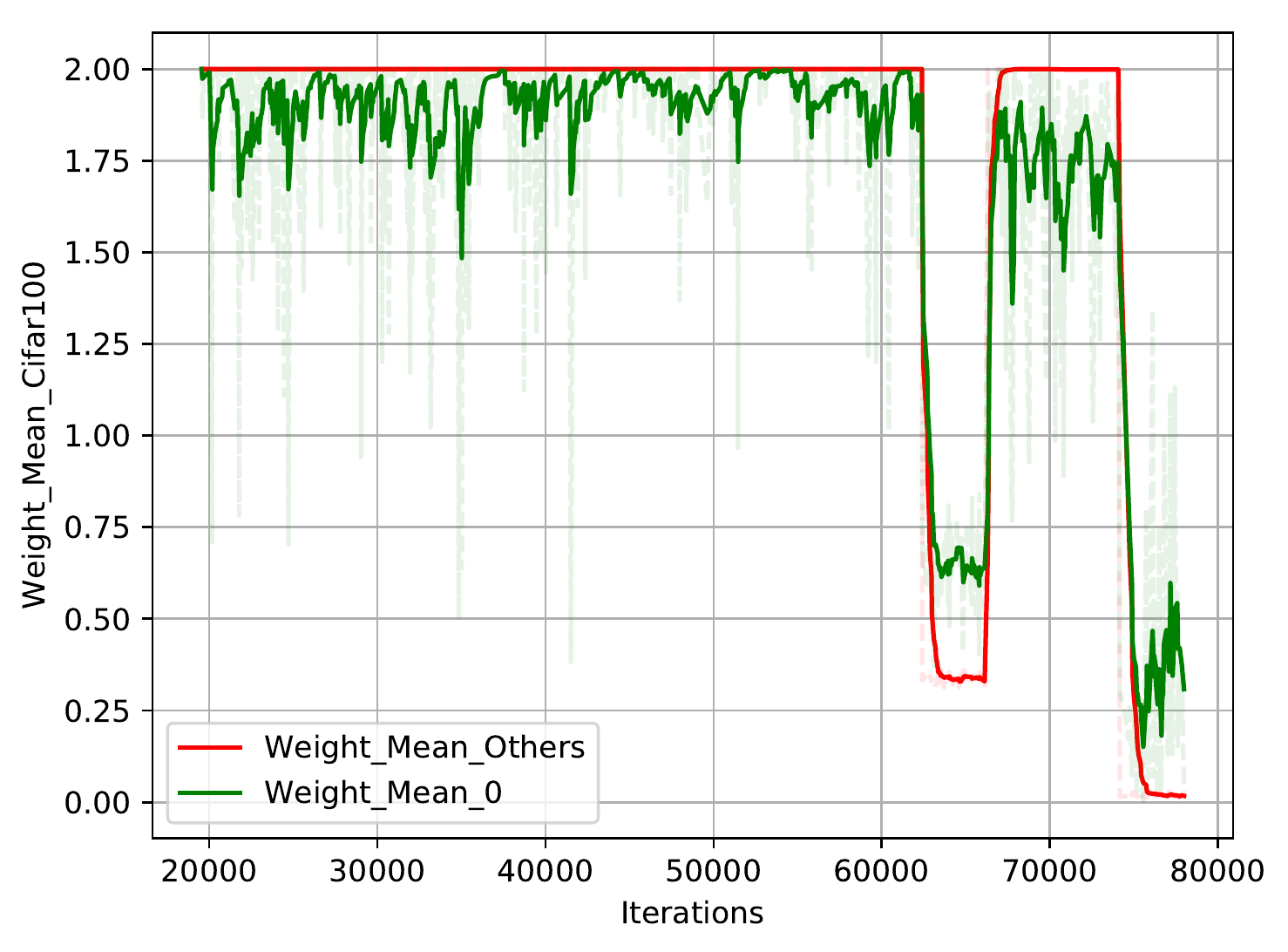}
			\caption{CIFAR100-Imbalance}
			\label{imbalance_weight_cifar100}
		\end{subfigure}
	\end{center}
	\caption{The weight means of classes with a small number of images (Weight Mean 0) and other classes with an original number of images (Weight Mean Others) on CIFAR.}
	\label{imbalace_weight_mean}
\end{figure}

On imbalance datasets,
we calculate the weight mean of samples from classes with a small number of images and
the weight mean of other classes in one batch along with training steps.
As showed in Figure~\ref{imbalance_weight_cifar10} and \ref{imbalance_weight_cifar100},
on both datasets,
the weights of the samples of label 0 is obvious larger than the samples of other labels.
What interesting is that in early steps,
the weights of small classes and other classes are close to each other,
but in the later stage, the difference of two weight becomes larger.
Our LAW can find samples of small classes and increase their weights.
The policy explored by LAW can deal with imbalance problems well.

%------------------------------------------------------------------------
\section{Conclusion}
In this paper,
we propose a novel example weighting framework called LAW,
which can learn weighting policy from data adaptively.
Experimental results demonstrate the superiority of weighting policy explored by LAW over standard training pipeline.

\bibliographystyle{aaai}
\bibliography{AAAI-LiZ.6749}

\begin{thebibliography}{}

\bibitem[\protect\citeauthoryear{Alain \bgroup et al\mbox.\egroup
  }{2015}]{alain2015variance}
Alain, G.; Lamb, A.; Sankar, C.; Courville, A.; and Bengio, Y.
\newblock 2015.
\newblock Variance reduction in sgd by distributed importance sampling.
\newblock {\em arXiv preprint arXiv:1511.06481}.

\bibitem[\protect\citeauthoryear{Arpit \bgroup et al\mbox.\egroup
  }{2017}]{arpit2017closer}
Arpit, D.; Jastrzebski, S.; Ballas, N.; Krueger, D.; Bengio, E.; Kanwal, M.~S.;
  Maharaj, T.; Fischer, A.; Courville, A.; Bengio, Y.; et~al.
\newblock 2017.
\newblock A closer look at memorization in deep networks.
\newblock {\em arXiv preprint arXiv:1706.05394}.

\bibitem[\protect\citeauthoryear{Bengio \bgroup et al\mbox.\egroup
  }{2009}]{bengio2009curriculum}
Bengio, Y.; Louradour, J.; Collobert, R.; and Weston, J.
\newblock 2009.
\newblock Curriculum learning.
\newblock In {\em Proceedings of the 26th annual international conference on
  machine learning},  41--48.
\newblock ACM.

\bibitem[\protect\citeauthoryear{Cui \bgroup et al\mbox.\egroup
  }{2019}]{cui2019class}
Cui, Y.; Jia, M.; Lin, T.-Y.; Song, Y.; and Belongie, S.
\newblock 2019.
\newblock Class-balanced loss based on effective number of samples.
\newblock In {\em Proceedings of the IEEE Conference on Computer Vision and
  Pattern Recognition},  9268--9277.

\bibitem[\protect\citeauthoryear{Deng \bgroup et al\mbox.\egroup
  }{2009}]{deng2009imagenet}
Deng, J.; Dong, W.; Socher, R.; Li, L.-J.; Li, K.; and Fei-Fei, L.
\newblock 2009.
\newblock Imagenet: A large-scale hierarchical image database.
\newblock In {\em 2009 IEEE conference on computer vision and pattern
  recognition},  248--255.
\newblock Ieee.

\bibitem[\protect\citeauthoryear{Graves \bgroup et al\mbox.\egroup
  }{2017}]{graves2017automated}
Graves, A.; Bellemare, M.~G.; Menick, J.; Munos, R.; and Kavukcuoglu, K.
\newblock 2017.
\newblock Automated curriculum learning for neural networks.
\newblock In {\em Proceedings of the 34th International Conference on Machine
  Learning-Volume 70},  1311--1320.
\newblock JMLR. org.

\bibitem[\protect\citeauthoryear{Han \bgroup et al\mbox.\egroup
  }{2018}]{han2018co}
Han, B.; Yao, Q.; Yu, X.; Niu, G.; Xu, M.; Hu, W.; Tsang, I.; and Sugiyama, M.
\newblock 2018.
\newblock Co-teaching: Robust training of deep neural networks with extremely
  noisy labels.
\newblock In {\em NIPS},  8527--8537.

\bibitem[\protect\citeauthoryear{He \bgroup et al\mbox.\egroup
  }{2016}]{he2016deep}
He, K.; Zhang, X.; Ren, S.; and Sun, J.
\newblock 2016.
\newblock Deep residual learning for image recognition.
\newblock In {\em Proceedings of the IEEE conference on computer vision and
  pattern recognition},  770--778.

\bibitem[\protect\citeauthoryear{Jiang \bgroup et al\mbox.\egroup
  }{2014a}]{jiang2014easy}
Jiang, L.; Meng, D.; Mitamura, T.; and Hauptmann, A.~G.
\newblock 2014a.
\newblock Easy samples first: Self-paced reranking for zero-example multimedia
  search.
\newblock In {\em Proceedings of the 22nd ACM international conference on
  Multimedia},  547--556.
\newblock ACM.

\bibitem[\protect\citeauthoryear{Jiang \bgroup et al\mbox.\egroup
  }{2014b}]{jiang2014self}
Jiang, L.; Meng, D.; Yu, S.-I.; Lan, Z.; Shan, S.; and Hauptmann, A.
\newblock 2014b.
\newblock Self-paced learning with diversity.
\newblock In {\em Advances in Neural Information Processing Systems},
  2078--2086.

\bibitem[\protect\citeauthoryear{Jiang \bgroup et al\mbox.\egroup
  }{2015}]{jiang2015self}
Jiang, L.; Meng, D.; Zhao, Q.; Shan, S.; and Hauptmann, A.~G.
\newblock 2015.
\newblock Self-paced curriculum learning.
\newblock In {\em Twenty-Ninth AAAI Conference on Artificial Intelligence}.

\bibitem[\protect\citeauthoryear{Jiang \bgroup et al\mbox.\egroup
  }{2017}]{jiang2017mentornet}
Jiang, L.; Zhou, Z.; Leung, T.; Li, L.-J.; and Fei-Fei, L.
\newblock 2017.
\newblock Mentornet: Learning data-driven curriculum for very deep neural
  networks on corrupted labels.
\newblock {\em arXiv preprint arXiv:1712.05055}.

\bibitem[\protect\citeauthoryear{Kahn and Marshall}{1953}]{kahn1953methods}
Kahn, H., and Marshall, A.~W.
\newblock 1953.
\newblock Methods of reducing sample size in monte carlo computations.
\newblock {\em Journal of the Operations Research Society of America}
  1(5):263--278.

\bibitem[\protect\citeauthoryear{Khan \bgroup et al\mbox.\egroup
  }{2018}]{khan2018cost}
Khan, S.~H.; Hayat, M.; Bennamoun, M.; Sohel, F.~A.; and Togneri, R.
\newblock 2018.
\newblock Cost-sensitive learning of deep feature representations from
  imbalanced data.
\newblock {\em IEEE transactions on neural networks and learning systems}
  29(8):3573--3587.

\bibitem[\protect\citeauthoryear{Kingma and Ba}{2014}]{kingma2014adam}
Kingma, D.~P., and Ba, J.
\newblock 2014.
\newblock Adam: A method for stochastic optimization.
\newblock {\em arXiv preprint arXiv:1412.6980}.

\bibitem[\protect\citeauthoryear{Krizhevsky and
  Hinton}{2009}]{krizhevsky2009learning}
Krizhevsky, A., and Hinton, G.
\newblock 2009.
\newblock Learning multiple layers of features from tiny images.
\newblock Technical report, Citeseer.

\bibitem[\protect\citeauthoryear{Kumar, Packer, and
  Koller}{2010}]{kumar2010self}
Kumar, M.~P.; Packer, B.; and Koller, D.
\newblock 2010.
\newblock Self-paced learning for latent variable models.
\newblock In {\em Advances in Neural Information Processing Systems},
  1189--1197.

\bibitem[\protect\citeauthoryear{Lee and Grauman}{2011}]{lee2011learning}
Lee, Y.~J., and Grauman, K.
\newblock 2011.
\newblock Learning the easy things first: Self-paced visual category discovery.
\newblock In {\em CVPR 2011},  1721--1728.
\newblock IEEE.

\bibitem[\protect\citeauthoryear{Lillicrap \bgroup et al\mbox.\egroup
  }{2015}]{lillicrap2015continuous}
Lillicrap, T.~P.; Hunt, J.~J.; Pritzel, A.; Heess, N.; Erez, T.; Tassa, Y.;
  Silver, D.; and Wierstra, D.
\newblock 2015.
\newblock Continuous control with deep reinforcement learning.
\newblock {\em arXiv preprint arXiv:1509.02971}.

\bibitem[\protect\citeauthoryear{Lin \bgroup et al\mbox.\egroup
  }{2017}]{lin2017focal}
Lin, T.-Y.; Goyal, P.; Girshick, R.; He, K.; and Doll{\'a}r, P.
\newblock 2017.
\newblock Focal loss for dense object detection.
\newblock In {\em Proceedings of the IEEE international conference on computer
  vision},  2980--2988.

\bibitem[\protect\citeauthoryear{Ma \bgroup et al\mbox.\egroup
  }{2017}]{ma2017self}
Ma, F.; Meng, D.; Xie, Q.; Li, Z.; and Dong, X.
\newblock 2017.
\newblock Self-paced co-training.
\newblock In {\em Proceedings of the 34th International Conference on Machine
  Learning-Volume 70},  2275--2284.
\newblock JMLR. org.

\bibitem[\protect\citeauthoryear{Neyshabur \bgroup et al\mbox.\egroup
  }{2017}]{neyshabur2017exploring}
Neyshabur, B.; Bhojanapalli, S.; McAllester, D.; and Srebro, N.
\newblock 2017.
\newblock Exploring generalization in deep learning.
\newblock In {\em Advances in Neural Information Processing Systems},
  5947--5956.

\bibitem[\protect\citeauthoryear{Peng, Li, and
  Wang}{2019}]{peng2019accelerating}
Peng, X.; Li, L.; and Wang, F.-Y.
\newblock 2019.
\newblock Accelerating minibatch stochastic gradient descent using typicality
  sampling.
\newblock {\em arXiv preprint arXiv:1903.04192}.

\bibitem[\protect\citeauthoryear{Ren \bgroup et al\mbox.\egroup
  }{2018}]{ren2018learning}
Ren, M.; Zeng, W.; Yang, B.; and Urtasun, R.
\newblock 2018.
\newblock Learning to reweight examples for robust deep learning.
\newblock In {\em International Conference on Machine Learning},  4331--4340.

\bibitem[\protect\citeauthoryear{Sandler \bgroup et al\mbox.\egroup
  }{2018}]{sandler2018mobilenetv2}
Sandler, M.; Howard, A.; Zhu, M.; Zhmoginov, A.; and Chen, L.-C.
\newblock 2018.
\newblock Mobilenetv2: Inverted residuals and linear bottlenecks.
\newblock In {\em Proceedings of the IEEE Conference on Computer Vision and
  Pattern Recognition},  4510--4520.

\bibitem[\protect\citeauthoryear{Simonyan and
  Zisserman}{2014}]{simonyan2014very}
Simonyan, K., and Zisserman, A.
\newblock 2014.
\newblock Very deep convolutional networks for large-scale image recognition.
\newblock {\em arXiv preprint arXiv:1409.1556}.

\bibitem[\protect\citeauthoryear{Supancic and Ramanan}{2013}]{supancic2013self}
Supancic, J.~S., and Ramanan, D.
\newblock 2013.
\newblock Self-paced learning for long-term tracking.
\newblock In {\em Proceedings of the IEEE conference on computer vision and
  pattern recognition},  2379--2386.

\bibitem[\protect\citeauthoryear{Zagoruyko and
  Komodakis}{2016}]{zagoruyko2016wide}
Zagoruyko, S., and Komodakis, N.
\newblock 2016.
\newblock Wide residual networks.
\newblock {\em arXiv preprint arXiv:1605.07146}.

\bibitem[\protect\citeauthoryear{Zhang \bgroup et al\mbox.\egroup
  }{2016}]{zhang2016understanding}
Zhang, C.; Bengio, S.; Hardt, M.; Recht, B.; and Vinyals, O.
\newblock 2016.
\newblock Understanding deep learning requires rethinking generalization.
\newblock {\em arXiv preprint arXiv:1611.03530}.

\end{thebibliography}

\end{document}